\documentclass{article} 
\usepackage{iclr2025_conference,times}

\usepackage{amsmath,amsfonts,bm}

\def\eqref#1{equation~\ref{#1}}

\def\1{\bm{1}}

\def\va{{\bm{a}}}
\def\vb{{\bm{b}}}

\def\vk{{\bm{k}}}

\def\vo{{\bm{o}}}

\def\vq{{\bm{q}}}
\def\vr{{\bm{r}}}

\def\vv{{\bm{v}}}

\def\mA{{\bm{A}}}

\def\mK{{\bm{K}}}
\def\mL{{\bm{L}}}
\def\mM{{\bm{M}}}

\def\mO{{\bm{O}}}

\def\mQ{{\bm{Q}}}

\def\mV{{\bm{V}}}

\def\mZ{{\bm{Z}}}

\DeclareMathAlphabet{\mathsfit}{\encodingdefault}{\sfdefault}{m}{sl}
\SetMathAlphabet{\mathsfit}{bold}{\encodingdefault}{\sfdefault}{bx}{n}

\usepackage{hyperref}
\usepackage{url}
\usepackage{graphicx}
\usepackage{caption}
\usepackage{subcaption}
\usepackage[utf8]{inputenc} 
\usepackage[T1]{fontenc}    
\usepackage{hyperref}       
\usepackage{xcolor}
\usepackage{tikz}
\usepackage{multirow}
\usepackage{adjustbox}
\usepackage{lscape}
\usepackage{pifont}
\newcommand{\cmark}{\ding{51}}%
\newcommand{\xmark}{\ding{55}}%
\hypersetup{
    colorlinks,
    linkcolor={red!50!black},
    citecolor={blue!50!black},
    urlcolor={blue!80!black}
}
\usepackage{tikz}

\definecolor{darkred}{RGB}{166,28,0}

\usepackage{booktabs}       
\usepackage{amsfonts}       
\usepackage{amsmath}
\usepackage{amssymb}
\usepackage{nicefrac}       
\usepackage{microtype}      
\usepackage{xcolor}         
\usepackage{graphicx}
\usepackage{graphbox}
\usepackage{pgf}
\usepackage{algorithm}
\usepackage{algorithmic}
\usepackage{wrapfig}

\usepackage{fontawesome}
\title{Scaling Stick-Breaking Attention: \\An Efficient Implementation \\and In-depth Study}

\author{Shawn Tan \\
MIT-IBM Watson AI Lab \\
\texttt{shawntan@ibm.com}
\And
Songlin Yang \\
MIT \\
\texttt{yangsl66@mit.edu}\hspace{2em} \\
\AND
Aaron Courville \\
Mila, Université de Montréal \\
\texttt{courvila@mila.quebec} \\
\And
Rameswar Panda \\
MIT-IBM Watson AI Lab \\
\texttt{rpanda@ibm.com} \\
\And
Yikang Shen \\
MIT-IBM Watson AI Lab \\
\texttt{yikang.shen@ibm.com}
}

\newif\ificlrfinal
\iclrfinalcopy 

\begin{document}

\newcommand{\dhead}{{d_\mathrm{head}}} 
\maketitle
\renewcommand\ttdefault{cmvtt}
\begin{abstract}
The self-attention mechanism traditionally relies on the softmax operator, necessitating positional embeddings like RoPE, or position biases to account for token order.
But current methods using still face length generalisation challenges.
We investigate an alternative attention mechanism based on the stick-breaking process in larger scale settings.
The method works as follows: For each token before the current, we determine a break point, which represents the proportion of the stick, the weight of the attention, to allocate to the current token.
We repeat this on the remaining stick, until all tokens are allocated a weight, resulting in a sequence of attention weights.
This process naturally incorporates recency bias, which has linguistic motivations for grammar parsing \citep{shen2017neural}.
We study the implications of replacing the conventional softmax-based attention mechanism with stick-breaking attention.
We then discuss implementation of numerically stable stick-breaking attention and adapt Flash Attention to accommodate this mechanism.
When used as a drop-in replacement for current softmax+RoPE attention systems, we find that stick-breaking attention performs competitively with current methods on length generalisation and downstream tasks.
Stick-breaking also performs well at length generalisation, allowing a model trained with $2^{11}$ context window to perform well at $2^{14}$ with perplexity improvements.
\ificlrfinal
    \begin{center}
     \faGithubSquare~ \url{https://github.com/shawntan/stickbreaking-attention}
    \end{center}
\fi
\end{abstract}

\section{Introduction}
The Transformer architecture \citep{vaswani2017attention}  uses a self-attention mechanism based on the softmax operator that enables the model to weigh the importance of different tokens in the input data.
However, the reliance on softmax requires using positional embeddings to introduce information about the order of tokens, as the attention mechanism itself is permutation-invariant.
The sinusoidal position embedding as proposed in \citet{vaswani2017attention} has since evolved to relative positional embeddings  \citep{shaw2018self}.
Learned relative positional biases were used in the T5 model \citep{raffel2020exploring}, and later fixed relative positional biases in \citet{press2021train}. 
At the time of writing, a commonly used form of position embedding is RoPE \citep{su2021roformer}. 
\cite{allen2023physics} observe that, in a context-free grammar parsing setting, attention mechanisms attend to the ``most adjacent'' non-terminal. 
This suggests an inclination to attend to the most recent entry that matches a given criteria. 
However, even with relative position information, it is possible to overfit on specific relative positions, resulting in failure to generalise.
\citet{kazemnejad2024impact} show that decoder-only Transformers with No Positional Embeddings (NoPE) can  implicitly recover positional information, experimental results suggest that NoPE Transformers generalise better on length.
While this is promising, a higher attention score from an irrelevant token in the sequence can function as a distractor \citep{kazemnejad2024impact,xiao2023efficient}.

The stick-breaking process may have properties that can alleviate the previously mentioned issues, and possess the `most recent' bias from \cite{allen2023physics} that we want.
For a token at position $j$ attending to position $i$, suppose the attention weight is given by: 
$$\mA_{i,j} =  \beta_{i,j} \prod_{i<k<j} \left( 1 - \beta_{k,j} \right) = \sigma(z_{i,j})  \prod_{i<k<j} \left( 1 - \sigma\left(z_{k,j}\right)\right),$$
where $z_{k,j}$ are the attention logits.
To illustrate via intuition, for $\mA_{i,j}$ to be high, all $\beta_{k,j}$ for $i < k$ have to be low.
Conversely, as long as any $\beta_{k,j}$ for $i < k$  is high, $\mA_{i,j}$ will be low, as a token between $i$ and $j$ has already been attended to. 
\begin{figure}
    \centering
    \vspace{-1em}
    \resizebox{0.55\linewidth}{!}{%
    \renewcommand{\arraystretch}{2.5}
    \vspace{-.5em}
    \begin{tabular}{rl}
        Logits & $z_{i,j} = \frac{\displaystyle\vq_j^\top \vk_i}{\displaystyle\sqrt{\dhead}}$ \\[1.2em]
        Softmax & $\displaystyle \mA_{i,j} = \displaystyle\frac{\exp(z_{i,j})}{\sum_{k=1}^j \exp(z_{k,j})}$ \\[1.2em]
        Stick-breaking & 
         $\displaystyle \mA_{i,j} = \sigma(z_{i,j})  \prod_{i<k<j} \left( 1 - \sigma(z_{k,j})\right)$
    \end{tabular}
    }\includegraphics[width=0.45\linewidth,align=c]{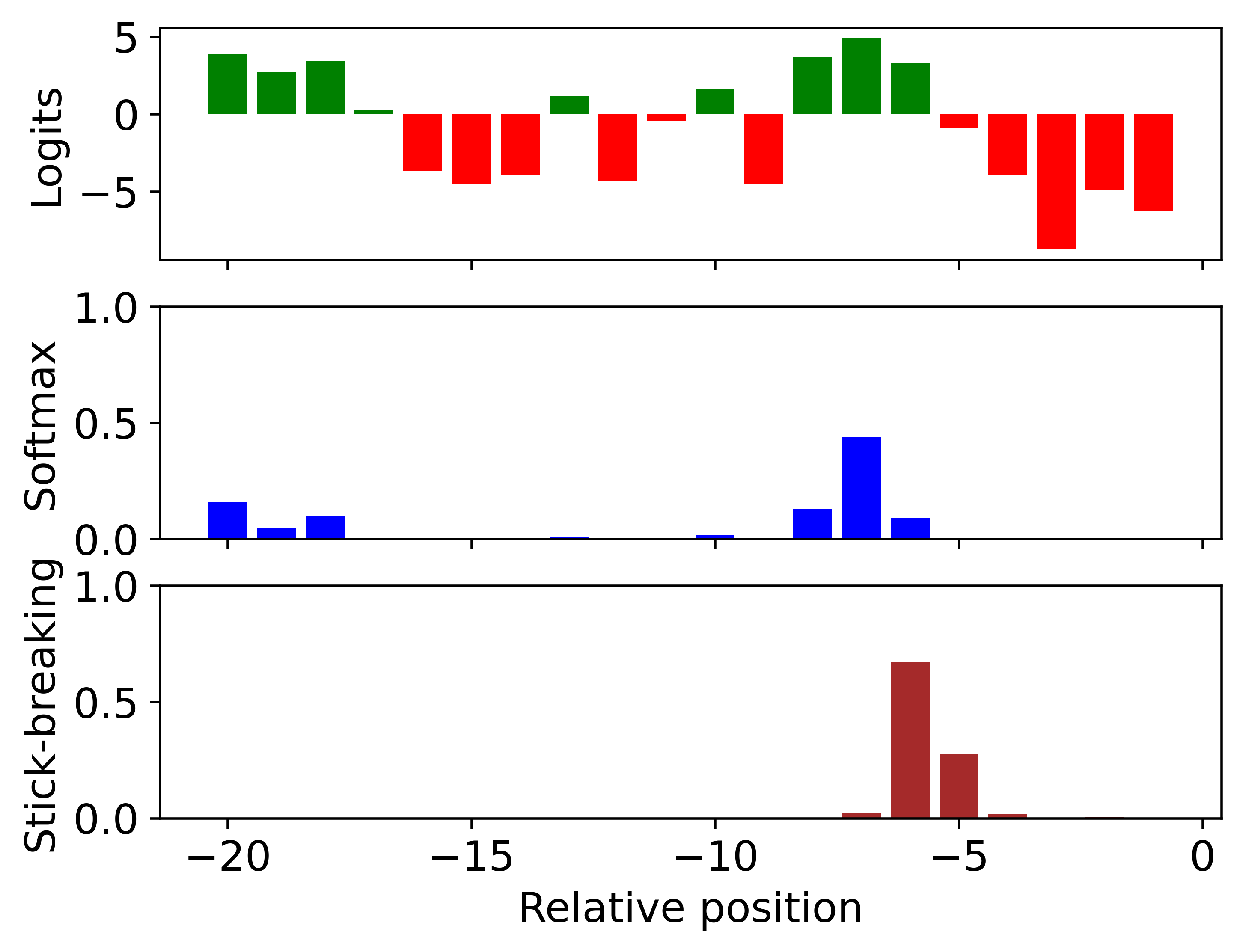}

    \caption{Differences in formulation between stick-breaking and softmax. Stick-breaking assigns high weights to the most recent high logit, while softmax will assign equal weightage to equal logits. 
    $\sigma(\cdot)$ can be any function $\mathbb{R} \rightarrow (0,1)$. In this paper we use $\sigma(x) = \frac{1}{1+exp(-x)}$}
    \label{fig:stickbreaking}
\end{figure}
\citet{shen2017neural} makes a similar observation as in \citet{allen2023physics}, and explicitly uses a stick-breaking process to model local structure. 
\citet{csordas2021neural} has a similar construction which they call Geometric attention after the Geometric distribution.
Geometric distribution only has one parameter $p$, which gives the probability of success per trial. 
The geometric distribution then gives the probability for which $k$ trials are needed for the first success: $(1 - p)^{k-1}p$.
But in  both stick-breaking and in Geometric attention \citep{csordas2021neural}, each $p$ is assigned a different value that corresponds to the attention score between two tokens.

In this paper, we expand upon prior work on this attention mechanism \citep{csordas2021neural,shen2023moduleformer}.
We focus on the implications of replacing the softmax-based attention mechanism with the stick-breaking process:
\begin{enumerate}
    \item We compare the different properties of stick-breaking attention against softmax attention,
    \item  We discuss numerically stable implementations of the stick-breaking attention, and make stick-breaking amenable for large-scale training by implementing a kernel for stick-breaking in Triton,
    \item We show the performance of stick-breaking attention on length-generalisation in language modelling, and evaluate 1B and 3B parameter models on various NLP tasks.
\end{enumerate}

\section{Stick-breaking Attention}

For a sequence of $L$ tokens, we have query $\vq_i \in \mathbb{R}^\dhead$, key $\vk_i \in \mathbb{R}^\dhead$, and value $\vv_i \in \mathbb{R}^\dhead$ vectors for $1 \leq i \leq L$. 
Then the attention weight for token at $j$ attending to position $i$ is computed by:
\begin{align}
\vo_{j} &= \sum_{i=1}^{j-1} \mA_{i,j} \vv_i, &
  \mA_{i,j} &= \beta_{i,j} \prod_{i<k<j} \left( 1 - \beta_{k,j} \right),  &
  \beta_{i,j} &= \sigma\left( z_{i,j} \right), &
  z_{i,j} &= \frac{\vq_j^\top \vk_i}{\sqrt{\dhead}}  \label{eqn:stickbreaking}
\end{align}
Equation \ref{eqn:stickbreaking} is the main difference between our proposal and softmax attention.
As discussed earlier and in \citet{csordas2021neural}, this parameterisation biases towards recency.
Specifically, for any pair of $i$ and $i'$ such that  $|j - i| < |j - i'|$ and $z_{i, j} = z_{i', j}$,  then $\mA_{i, j} \geq \mA_{i',j}$. 
Consequently, this imposes an ordering on the attention, and we do not use position embeddings with the query and key embeddings.

We consider two sets of logits, $z_{i,j}$ and $z'_{i,j}$ and their respective attention weights  $\mA_{i,j}$ and $\mA'_{i,j}$.
If $z_{i,k} = z'_{i,k}$ for $i < k < j$, then $\mA_{i,j} = \mA'_{i,j}$.
This means that unlike softmax attention, a high attention score further back in the sequence does not `distract' from a more recent high $z_{i,j}$ score.
Further, if $\sum_{k=i}^{j-1} \mA_{k,j} = 1$, then the output is invariant to appending additional context earlier than $i$.
We note that $\sum_{i=1}^{j-1} \mA_{i,j} \leq 1$, which allows this attention mechanism to attend to nothing when all $\beta_{i,j} = 0$.
We discuss a strategy to deal with the remaining attention weight in Appendix \ref{sec:remainder}.

\section{Related Work}
\paragraph{Stick-breaking Process}
The stick-breaking process formulation of the Dirichlet process \citep{sethuraman1994constructive} is also known as the GEM distribution, first coined in \citet{ewens1990population} after \citet{griffiths1989genealogical}, \citet{engen1975note}, and \citet{mccloskey1965model}.
The GEM is a specific case of what was known as a Residual Allocation Model (RAM; \citealt{allen1976environmental}).
There are instances of the distribution being used as a differentiable attention-like mechanism in neural models. 
\citet{shen2017neural} used stick-breaking process for modelling language, and showed that the model can induce grammatical structure to some extent. 
\citet{csordas2021neural} used stick-breaking attention, which they refer to as Geometric attention, in a bidirectional encoder  set up. 
\citet{shen2023moduleformer} used stick-breaking attention in a decoding-only setup, but does not explicitly study the properties of stick-breaking.

\paragraph{Softmax attention, Positional embeddings, and Length Generalisation}
\citet{bondarenko2024quantizable} observe that softmax attention tends to attend to low-information tokens with high scores in order to `do nothing'.
\citet{xiao2023efficient} introduces \emph{attention sinks}, a learnable token that the attention can assign attention weights to.
\citet{irie2019language} and \citet{haviv2022transformer} find that in a decoder-only setting, a Transformer with no positional embedding can work fairly well. 
\citet{kazemnejad2024impact} also found similar results, while also showing that NoPE has a tendency to attend to the start of the sequence, while ALiBi \citep{press2021train} has a tendency to only attend to the most recent tokens. 
However, \citet{zhou2024transformers} later found that Transformers without position embeddings do not generalise to out-of-distribution sequence lengths for an addition task. 
At present,  Rotary Positional Embeddings (RoPE; \citealt{su2021roformer}) are the most commonly used position embedding.
It encodes relative positions via multiplicative interactions with the key and query. 
RoPE has been found to generalise to out-of-distribution lengths poorly \citep{press2021train,zhou2024transformers,kazemnejad2024impact}, but a common trick to extend the context window RoPE-based Transfomers is to use NTK-aware RoPE scaling \citep{blocntkaware}.
\paragraph{Conditional Computation}
The use of stick-breaking for conditional computation has also been explored. 
\citet{tan2016towards} uses a the stick-breaking distribution as a mixture over outputs for each layer in an MLP for an acoustic model.
\citet{graves2016adaptive}  also suggested a similar formulation for language modelling. 
Later,  \citet{banino2021pondernet} and \citet{tan2023sparse} also use a stick-breaking formulation for dynamic depth modelling in a Transformer model. 
These prior works use conditional computation on the \emph{depth} of the model, while in our case, we use stick-breaking as a method of restricting the computation \emph{length}-wise. 

\newcommand{\dhidden}{d_\mathrm{hidden}}
\paragraph{Connection to Selective State-space Models} Each stick-breaking attention head at every time-step can be viewed as the hidden state of the final step of a selective State-space Model (SSM;\citealt{gu2023mamba}). 
For a given time-step $j$, consider the following SSM in its recurrent form and its corresponding convolutional form (as described in \citealt{merrill2024illusion}):
\begin{align}
    \textrm{Let} \qquad \hat{\vo}_{i,j} &= (1 - \beta_{i,j}) \cdot \hat{\vo}_{i-1,j}  + \beta_{i,j} \cdot \vv_{j}, \\
     \textrm{then}\qquad \vo_j &=  \hat{\vo}_{j-1,j}  = \sum_{1 \leq i \leq j} \left(\beta_{i,j} \prod_{i < k < j} (1 - \beta_{k,j}) \right)\cdot\vv_i ,
\end{align}
which is equivalent to the first term in Equation \ref{eqn:stickbreaking}.
Intuitively, this means that the output head of stick-breaking attention is equivalent to the end state of a single-gate selective SSM.
Typically, an attention layer for a Transformer with hidden dimension $\dhidden$ has $h$ heads such that $\dhidden = h \cdot \dhead$. 
For equivalence with an SSM, we need a constant query vector and each dimension as a separate head, \textit{e.g.} $\vq_i = \mathbf{1}$ for all $i$, and $h=\dhidden, \dhead  = 1$.

\paragraph{Connection to Additive Relative Position Encoding (Additive RPE; \citealt{kazemnejad2024impact})} 
Generally, Additive RPEs incorporate an added bias function $g$ of the distance of the tokens $i - j$ and the maximum length of the sequence $L$:
\begin{align*}
    \mA_{ij} &\propto  \exp\left(\vq_j ^ \top \vk_i + b \right)
\end{align*}
In the case of ALiBi \citep{press2021train}, this is a linear function $b  = -m \cdot (j - i)$.
This implies that the attention weights will drop off exponentially the further $j$ and $i$ are apart, regardless of the attention scores.
In stick-breaking, Equation \ref{eqn:forward} has a form that accounts for the scores from $j$ to $i$ with the bias $b = - \sum_{k=i+1}^{j}  \log \left(1 + \exp(z_{k,j})\right)$.
Specifically, if $ \log \left(1 + \exp(z_{k,j})\right) \geq m$, then $b \leq -m\cdot(j - i)$.
This suggests a learnable relative position bias that is dependent on the intermediate scores between $j$ and $i$.

\section{Implementation}

Implementing stick-breaking attention naively in PyTorch results in realising the $L^2$ matrix for the attention logits (where $L$ is the length of the input).
FlashAttention \citep{dao2022flashattention} reduces the memory footprint of attention by side-stepping the $O(L^2)$ memory complexity of realising the attention matrix. 
In order to achieve this, it only realises blocks of the attention weights during computation, and accumulates the resulting weighted sum of $\vv_i$.
We take a similar approach to speeding up stick-breaking attention, allowing it to be used for longer sequences.

\paragraph{Forward}
\newcommand*\circled[1]{\tikz[baseline=(char.base)]{
            \node[shape=circle,draw,inner sep=2pt] (char) {#1};}}

Computing Equation \ref{eqn:stickbreaking} directly will result in underflow issues, especially with lower precision training.
We perform the operations in log-space, which results in a cumulative sum instead:
\begin{align}
    \mA_{i,j}= \exp\left( \log \beta_{i,j} + \sum_{k=i+1}^{j-1} \log \left( 1 - \beta_{k,j} \right)   \right) 
    = \exp\left( z_{i,j} - \sum_{k=i}^{j-1}  \log \left(1 + \exp(z_{k,j})\right)   \right)  \label{eqn:forward}
\end{align}
Where $\log\left(1 + \exp(\cdot)\right)$ is commonly known as the $\mathrm{softplus}$ operation. See Appendix \ref{eqn:logspacederiv} for the full derivation.
We further numerically stabilise $\mathrm{softplus}$ with the following computation:
\begin{equation}
    \mathrm{softplus}(x) = 
\begin{cases}
    \log\left( 1 + \exp(x)\right), & \text{if } x\leq 15\\
    x              & \text{otherwise}
\end{cases}\label{eqn:softplus}
\end{equation}
to prevent overflowing of $\exp(x)$.
To further speed up the computation of $\mathrm{softplus}$, we also write the operation in Parallel Thread Execution (PTX).

\paragraph{Backward}

Let $\tilde{\mA}_{i,j} = \log \mA_{i,j}$, then: 
\begin{align}
    \dfrac{\partial \mathcal{L} }{\partial \tilde{\mA_{i,j}}} &= \dfrac{\partial \mathcal{L} }{\partial\mA_{i,j}} \cdot \mA_{i,j}, &
    \frac{\partial \mathcal{L}}{\partial z_{i,j}} &= 
    \underbrace{\dfrac{\partial \mathcal{L} }{\partial \tilde{\mA_{i,j}}}}_{\text{Contribution from }i,j}
    - \underbrace{\sigma(z_{i,j}) \sum_{i'=1}^{j-1}  \dfrac{\partial \mathcal{L} }{\partial \tilde{\mA_{i',j}}}}_{\text{Contribution from before }i, j} \label{eqn:backward}
\end{align}

The above equations dictate the direction of the order of computation for our implementation.
For the forward pass (Eqn. \ref{eqn:forward}), we compute from $j$ to $1$, backwards through time and accumulate $ \sum_{k=i+1}^{j}  \log \left(1 + \exp(z_{k,j})\right)$.
For backward pass (Eqn. \ref{eqn:backward}), we compute from $1$ to $j$, accumulating  $\sum_{j'=1}^{j-1}  \frac{\partial \mathcal{L} }{\partial \tilde{\mA_{i,j'}}}$.


\subsection{Triton Specifics}
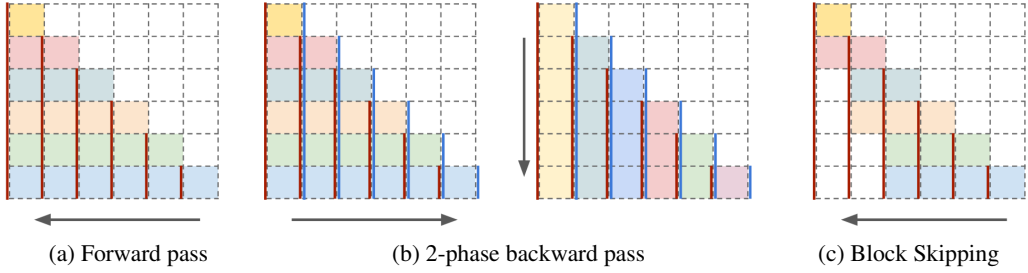
\begin{figure}
    \centering
\begin{subfigure}{0.33\linewidth}
\centering
\begin{tikzpicture}[scale=0.54]
    \fill[yellow!20] (0,5) rectangle (1,6);
    \fill[red!10] (0,4) rectangle (2,5);
    \fill[blue!10] (0,3) rectangle (3,4);
    \fill[orange!20] (0,2) rectangle (4,3);
    \fill[green!20] (0,1) rectangle (5,2);
    \fill[cyan!20] (0,0) rectangle (6,1);
    \draw [gray, dashed, line width=1.5pt] (0,0) grid  (6,6);
    \node[right] at (6,3) {$\mQ$};
    \node[above] at (3,6) {$\mK, \mV$};
    \node[left] at (0,3) {$\mO$};
    \draw [darkred, thick, line width=2pt] (0,0) -- (0,6);
    \draw [thick, <-, line width=1.5pt] (1,-0.5) -- (5,-0.5)  node[midway, below] {Accumulate $\mO$};
    \draw [white!0, thick, -, line width=0pt] (1,-1.5) -- (5,-1.5);
\end{tikzpicture}
\caption{Forward pass}\label{fig:forward_grid}
\end{subfigure}
\begin{subfigure}{0.33\linewidth}
\centering
\begin{tikzpicture}[scale=0.54]
    \fill[yellow!20] (0,5) rectangle (1,6);
    \fill[red!10] (0,4) rectangle (2,5);
    \fill[blue!10] (0,3) rectangle (3,4);
    \fill[orange!20] (0,2) rectangle (4,3);
    \fill[green!20] (0,1) rectangle (5,2);
    \fill[cyan!20] (0,0) rectangle (6,1);
    \draw [gray, dashed, line width=1.5pt] (0,0) grid  (6,6);
    \node[right] at (6,3) {$\nabla\mQ$};
    \node[above] at (3,6) {$\nabla\mK, \nabla\mV$};
    \node[left] at (0,3) {$\nabla\mO$};
    \draw [darkred, thick, line width=2pt] (0,0) -- (0,6);
    \draw [thick, ->, line width=1.5pt] (1,-0.5) -- (5,-0.5)  node[midway, below] {Accumulate $\nabla \mQ$};
    \draw[blue,bend left, ->, line width=1pt] (1.5,0.5)  to[out=45, in=135] node[midway, right]{\textbf{atomic add}} (1.5,6.5);
    \draw [white!0, thick, -, line width=0pt] (1,-1.5) -- (5,-1.5);
\end{tikzpicture}
    \caption{Backward pass}\label{fig:backward_grid}
\end{subfigure}
\begin{subfigure}{0.32\linewidth}
\begin{center}
\begin{tikzpicture}[scale=0.54]
    \fill[yellow!20] (0,5) rectangle (1,6);
    \fill[red!10] (0,4) rectangle (2,5);
    \fill[blue!10] (1,3) rectangle (3,4);
    \fill[orange!20] (1,2) rectangle (4,3);
    \fill[green!20] (3,1) rectangle (5,2);
    \fill[cyan!20] (3,0) rectangle (6,1);
    \draw [gray, dashed, line width=1.5pt] (0,0) grid  (6,6);
    \node[right,white] at (6,3) {$\nabla\mQ$};
    \node[above,white] at (3,6) {$\nabla\mK, \nabla\mV$};
    \node[left,white] at (0,3) {$\nabla\mO$};
    \draw [white!0, thick, -, line width=0pt] (1,-1.5) -- (5,-1.5);
\end{tikzpicture}
    \caption{Block skipping}\label{fig:skip_grid}
    \end{center}
\end{subfigure}
    \caption{Thread tile assignments for a given attention head and a sequence. Tiles coloured the same are processed by the same thread.
For stick-breaking, the forward pass has to be computed from right-to-left, while the backward pass is computed in the reverese order. 
Uncoloured tiles are not computed: upper right tiles are not used in causal language modelling, and in the case of block skipping, some blocks can be skipped if all entries have summed to 1.}
    \label{fig:sweep_diagrams}
\end{figure}

We  modify the Triton implementation of Flash Attention for accelerating the stick-breaking attention.
We used a similar approach as in Flash Attention 2 (\citealt{dao2023flashattention}; FA2), loading blocks of $\mQ$ from High Bandwidth Memory (HBM) into Static random-access memory (SRAM).
In the forward pass, we accumulate towards the start of the sequence, loading blocks of $\mK$ and $\mV$ the values $\textcolor{darkred}{\va}$, finally writing $\textcolor{darkred}{\va}$ and $\mO$ as the output into HBM.
In the backward pass, we differ from the FA2 approach.
FA2 accumulates $\nabla\mK, \nabla\mV$, and makes atomic adds toward $\nabla\mQ$, which allows for fewer synchronisations during the computation of the backward pass. 
However, in the case of stick-breaking, the accumulation of the gradients have to be in the reverse direction of the cumulative sum.
As a result, we have to do atomic adds for both $\nabla\mK$ and $\nabla\mV$.
The resulting memory complexity of our approach is still $O(L)$, which allows us to train large models with stick-breaking.

In our implementation, the block size is $d_\mathrm{block} = 64$.
Algorithms \ref{alg:forward} details the forward pass (illustrated in Figure \ref{fig:forward_grid}), and Algorithm \ref{alg:backward} details the backward pass (illustrated in Figure \ref{fig:backward_grid}).

\newcommand{\blocksize}{d_\mathrm{block}}
\begin{algorithm}[t]
\small
\begin{algorithmic}
   \caption{\textsc{Forward} thread $i$} \label{alg:forward}
    \STATE $\textcolor{darkred}{\va} \leftarrow \mathbf{0}, \mO \leftarrow \mathbf{0}$  \hfill \textit{\small // init  $\textcolor{darkred}{\va}: \blocksize \times 1,\quad\mO:  \blocksize \times \dhead$}
    \STATE $\mQ \leftarrow $ load block $i$ of $\mQ$ 
    \hfill \textit{// load from HBM to SRAM, $\mQ:\blocksize \times \dhead$} 
    \FOR{$k$ in $i \hdots 1$}
 
        \STATE $\mK \leftarrow $ load block $k$ of $\mK$  \hfill \textit{// $\mK:\blocksize \times \dhead$}
        \STATE $\mV \leftarrow $ load block $k$ of $\mV$ 
        \STATE \vspace{-1.1em}~\hspace{10em}\rlap{\smash{$\left.\begin{array}{@{}c@{}}%
        {}\\ {}\\ {}\\ {}\\ \end{array}\color{orange}\right\}\hfill %
          \color{orange}\begin{tabular}{l}\circled{1}\end{tabular}$}}
        \hfill \textit{// $\mV:\blocksize \times \dhead$}
        \STATE $\mZ \leftarrow \mQ\mK^\top$  %
        \hfill \textit{// $\mZ:\blocksize \times \blocksize$}
        \STATE $\mL \leftarrow -\mathrm{softplus}(\mZ)$ \hfill\textit{// $\mL:\blocksize \times \blocksize$}
        \STATE {$\mA \leftarrow \exp\left( \mZ + \mathrm{cumsum}_{\leftarrow} \mL + \textcolor{darkred}{\va}\right)$}\hfill \textit{//  cumulative sum right to left,  $\mA:\blocksize \times \blocksize$}
        \STATE $\mO \leftarrow \mO + \mA\mV$
        \STATE $\textcolor{darkred}{\va} \leftarrow \textcolor{darkred}{\va} + \sum_\leftarrow \mL$ 

    \ENDFOR
    \STATE $\textcolor{darkred}{\va} \rightarrow$ store block $i$ of $\textcolor{darkred}{\mM}$ \hfill \textit{// store from SRAM to HBM}
    \STATE $\mO \rightarrow$ store block $i$ of $\vo$
\end{algorithmic}
\end{algorithm}
\begin{algorithm}[t]
\small
\begin{algorithmic}
   \caption{\textsc{Backward} thread $i$} \label{alg:backward}
   \STATE $\vb \leftarrow \mathbf{0}$
    \hfill \textit{\small // init  $\vb: \blocksize \times 1$}
    \STATE $\nabla\mO \leftarrow $ load block $i$ of $\nabla\mO$
    \hfill \textit{\small //~$\nabla\mO:  \blocksize \times \dhead$}
    \STATE $\mQ \leftarrow $ load block $i$ of $\mQ$
    \hfill \textit{\small //~$\mQ:  \blocksize \times \dhead$}
    \STATE $\textcolor{darkred}{\va} \leftarrow$ load block $i$ of $\textcolor{darkred}{\mM}$
    \hfill \textit{\small //~$\textcolor{darkred}{\va}: \blocksize \times 1$}
    \FOR{$k$ in $1 \hdots i$}
        \STATE \textcolor{orange}{\textbf{Do } \circled{1}}
        \STATE $\textcolor{darkred}{\va} \leftarrow \textcolor{darkred}{\va} - \sum_\leftarrow \mL$ 
        \STATE {$\mA \leftarrow \exp\left( \mZ + \mathrm{cumsum}_{\leftarrow} \mL + \textcolor{darkred}{\va}\right)$}
        \STATE $\nabla \tilde{\mA} \leftarrow \mA \odot ( \nabla\mO \mV^\top)$
        \hfill \textit{// $\nabla \tilde{\mA}:\blocksize \times \blocksize$}
        \STATE $\nabla \mZ \leftarrow \nabla \tilde{\mA}  - (1 - \exp(\mL)) \odot \left(\mathrm{cumsum}_{\rightarrow}\nabla \tilde{\mA}  +  \vb \right) $
        \hfill \textit{// $\nabla \mZ:\blocksize \times \blocksize$}
        \STATE $\vb \leftarrow \vb + \sum_{\rightarrow} \nabla \tilde{\mA}$ 

        \STATE $\nabla\mQ \leftarrow \nabla\mQ + \nabla\mZ \mK^\top $
        \STATE $\mA^\top \nabla \mO \rightarrow \textrm{atomic add block $k$~} \nabla \mK$
        \hfill \textit{// accumulate gradient for $\nabla\mK$ and $\nabla \mV$}
        \STATE $\nabla \mZ^\top \mQ \rightarrow \textrm{atomic add block $k$~} \nabla \mV$
    \ENDFOR
    \STATE $\nabla\mQ  \rightarrow$ store block $i$ of $\nabla\mQ $
\end{algorithmic}
\end{algorithm}

\paragraph{Throughput} 
We measure throughput on Dolomite Engine \citep{Mishra_Dolomite_Engine_A_2024} with a 1B class model on a node with 8 H100 GPUs.
 Stick-breaking attains a throughput of 21 billion tokens / day. 
This is a 29\% performance drop compared to FA2, which attains a throughput of 29.5 billion tokens / day.
Despite the slowdown, the Triton implementation allows the method to be used for long sequences as the naive Torch implementation causes out-of-memory issues.
Further optimisation of the kernel may be possible if written in CUDA, taking advantage of synchronisation features the language exposes in the hardware.
\paragraph{Conditional computation}
We can incorporate speedups due to the specific nature of the stick-breaking process.
Since, we accumulate $\mA_{k,j} \cdot \vv_k$  from the diagonal to $k=1$,  when $\sum_{i=k}^{j-1} \mA_{i,j} = 1$  for some $k$, we know that all $\mA_{k',j} = 0$ where $k' < k$.
Therefore, once the condition is met, we can skip all subsequent accumulations. 
We implement block skipping only the forward pass, and measured the time elapsed on a subset of 10,000 instances of The Pile\footnote{\url{https://huggingface.co/datasets/NeelNanda/pile-10k}}.
Evaluating on 16K context lengths using the the LM evaluation harness \citep{eval-harness} with a batch size of 1, we find that early halting allows for a 9.3\% percent speed improvement.

\section{Experiments}
In this section, we compare existing attention methods against stick-breaking.
We first look at a modification of a synthetic task from \cite{arora2023zoology} to understand the inductive biases of stick-breaking.
We then compare the stick-breaking against existing length  extrapolation methods on a 350M model setting. 
We then pretrain a 1B parameter model, and evaluate it on various NLP benchmarks, and evaluate it for length extrapolation and retrieval capabilities on long context using the RULER benchmark \citep{hsieh2024ruler}.
We also report benchmark results for a 3B model we have trained.
For reference, the size of the models are detailed in Table \ref{tab:model_sizes}.

\subsection{Multi-Query Repeated Associative Recall Task}

To illustrate the inductive bias of stick-breaking attention, we first analyse its behaviour on a simple synthetic task. 
\citet{arora2023zoology} demonstrate that there is a correlation between \emph{multi-query associative recall} (MQAR) and the performance of language modelling. 
They show that while Transformers, which are based on attention can easily handle MQAR, the current linear state-space models cannot. 
\begin{wrapfigure}[12]{R}{0.4\textwidth}
    \centering
    \includegraphics[width=0.95\linewidth]{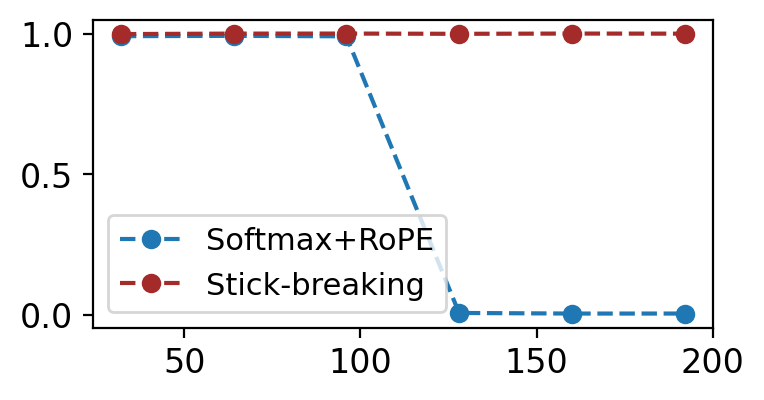}
    \caption{MQRAR performance on increasing key-value pairs.}
    \label{fig:mqrar}
\end{wrapfigure}
While stick-breaking can also solve the MQAR task, we formulated a different version of this toy task and tested it on both softmax attention and stick-breaking.
As before, each instance of the task has an initial assignment of values to variables.
However, in the query sequence, the same variable can be queried multiple times, and each query is then followed by a variable assignment, which successive queries must recall. 
We refer to this task as \emph{multi-query repeated associative recall} (MQRAR).
This setting can be analogous to some scenarios in programming where variables are repeatedly updated, and it where it is useful to understand the current state of the variable assignment.
The following is an example sequence of the task:
\begin{center}
    \begin{tabular}{rcccccccccccccccccc}
    \textbf{Input} &B & 6 & P & 4 & E & 3 & X & 1 & Z & 2 & E & 2 & B & 1 & E & 5 & B & 4\\
    \textbf{Output} & $\phi$ &$\phi$ &$\phi$ &$\phi$ &$\phi$ &$\phi$ &$\phi$ &$\phi$ &$\phi$ &$\phi$&
    3&$\phi$& 6& $\phi$&2&$\phi$&1 & $\phi$ 
    \end{tabular}
\end{center}

We compare Transformer models with Softmax and RoPE against stick-breaking attention, comparing their ability to handle MQRAR with increasing key-value pairs: from 32 to 192, in increments of 32.
The full sequence length is 768.
Both models are 2-layer Transformers with 256 hidden dimension and one attention head.
We sweep through 4 learning rates of $\{10^{-4}, 10^{-\frac{10}{3}}, 10^{-\frac{8}{3}}, 10^{-2}\}$, and report the results of the best performing model. 
Softmax+RoPE  is able to perform perform this task up to 128 key-value pairs, while stick-breaking is able to deal with sequences up to 192 key-value pairs.

\begin{figure}
    \centering
    \includegraphics[width=0.9\linewidth]{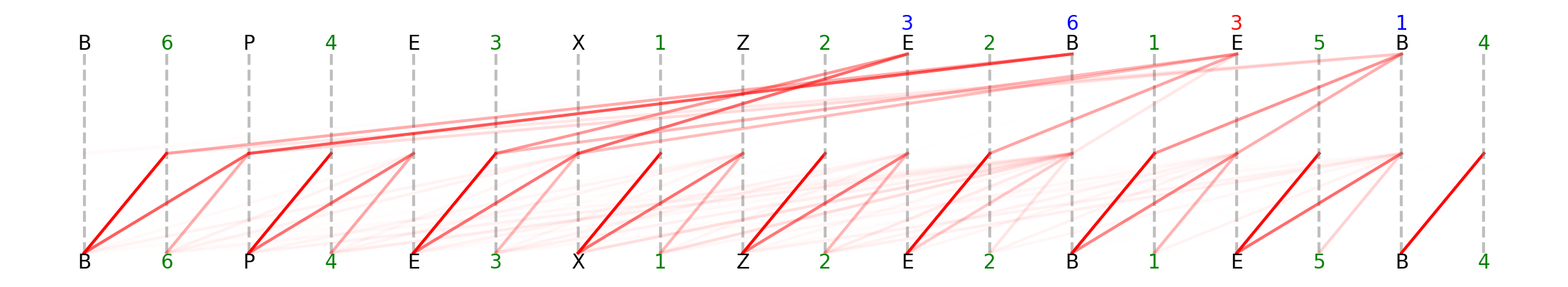}
    \includegraphics[width=0.9\linewidth]{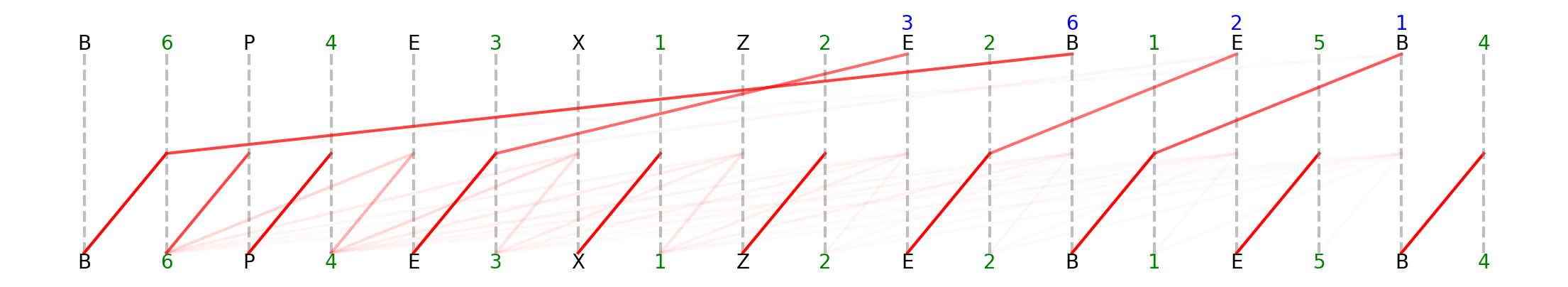}
    \caption{
    Attention visualisation of the models trained on the MQRAR task. The figure shows the attention for each token for the 2-layer 100-dimension Transformer.
    Note that in the standard Softmax+RoPE setting (above), the attention head is ``distracted" at the third retrieval of `E', attending to the first instance of `E' rather than the mroe recent one.
    In the stick-breaking setting (below), each attention head attends to the prior assignment of the variable. 
    }
    \label{fig:attention_visualisation}
\end{figure}
For a more qualitative analysis, we trained another set of 2-layer Transformers of 100 dimensions on 16 key-value pairs, and visualised the patterns of the attention head.
In Figure \ref{fig:attention_visualisation}, we visuallised the above given example.
Note here that the retrieval of the third `E' is distracted by the earlier assignment of 3 to `E', while stick-breaking attention correctly retrieves the later assignment of 2. 
This may be due to the limitations of RoPE with fewer layers and fewer head dimensions (here we use 32).


\subsection{350M Model Length Extrapolation}\label{sec:350extrapolation}
\begin{table}
\caption{Model hyperparameters and total size}
\label{tab:model_sizes}
    \centering
\begin{tabular}{lrrrrrr}
\toprule
 & $n_\mathrm{layer}$ & $\dhidden$ & $d_\mathrm{inter}$ & $n_\mathrm{head}$ & $L$ & \textbf{Total params.} \\
\midrule
350M & 24 & 1024 & 2730 & 32 & 2048 & 367,526,912 \\
1B & 40 & 1536 & 4096 & 24 & 4096 & 1,208,083,968 \\
3B & 40 & 2304 & 9216 & 36 & 4096 & 3,513,473,280 \\
\bottomrule
\end{tabular}

\end{table}
\begin{figure}[b]
    \centering
    \begin{subfigure}{0.315\linewidth}
    \includegraphics[width=\textwidth]{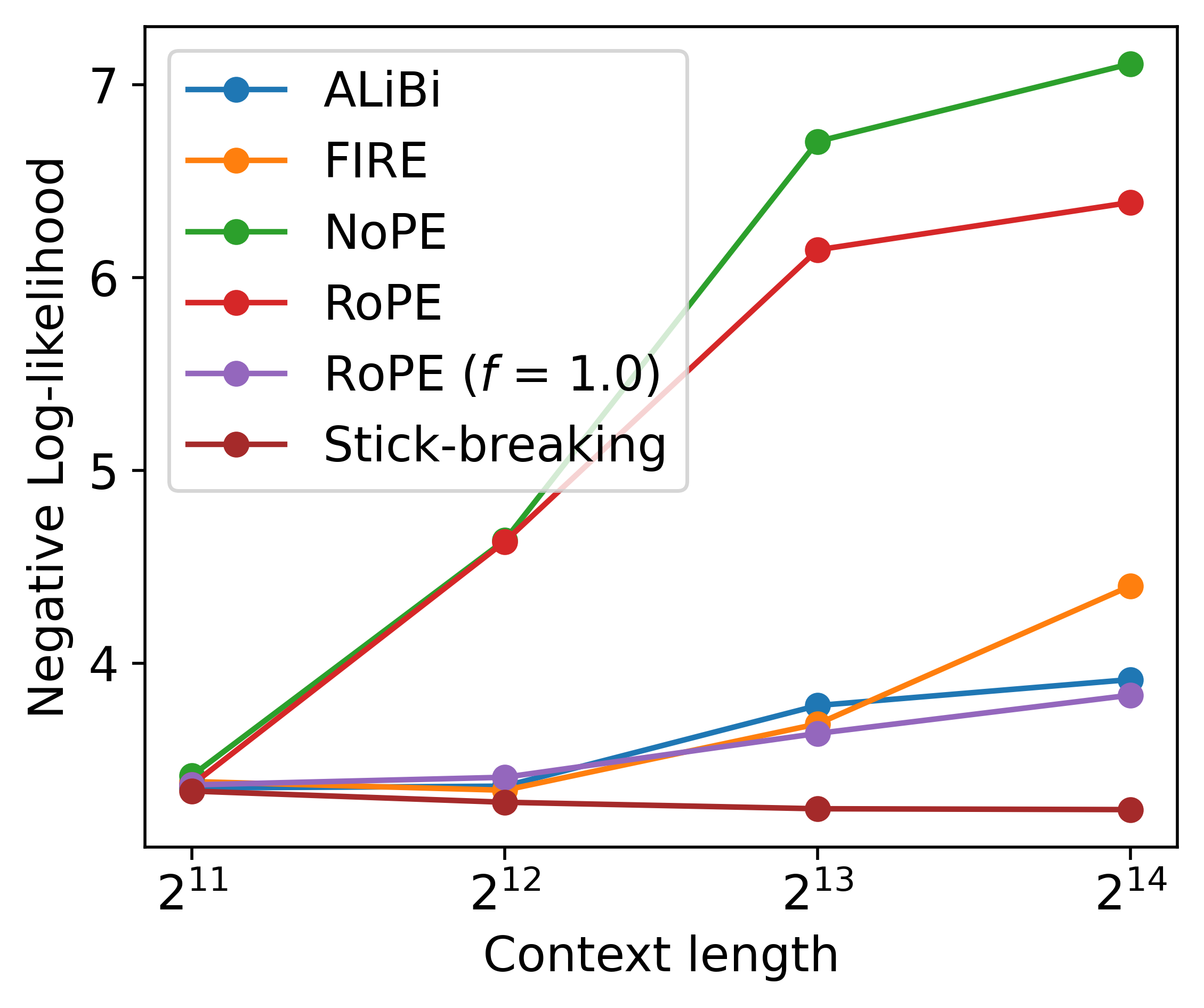}
    \caption{Position bias and embeddings}\label{fig:length_extrapolation}
\end{subfigure}
\begin{subfigure}{0.325\linewidth}
    \includegraphics[width=\textwidth]{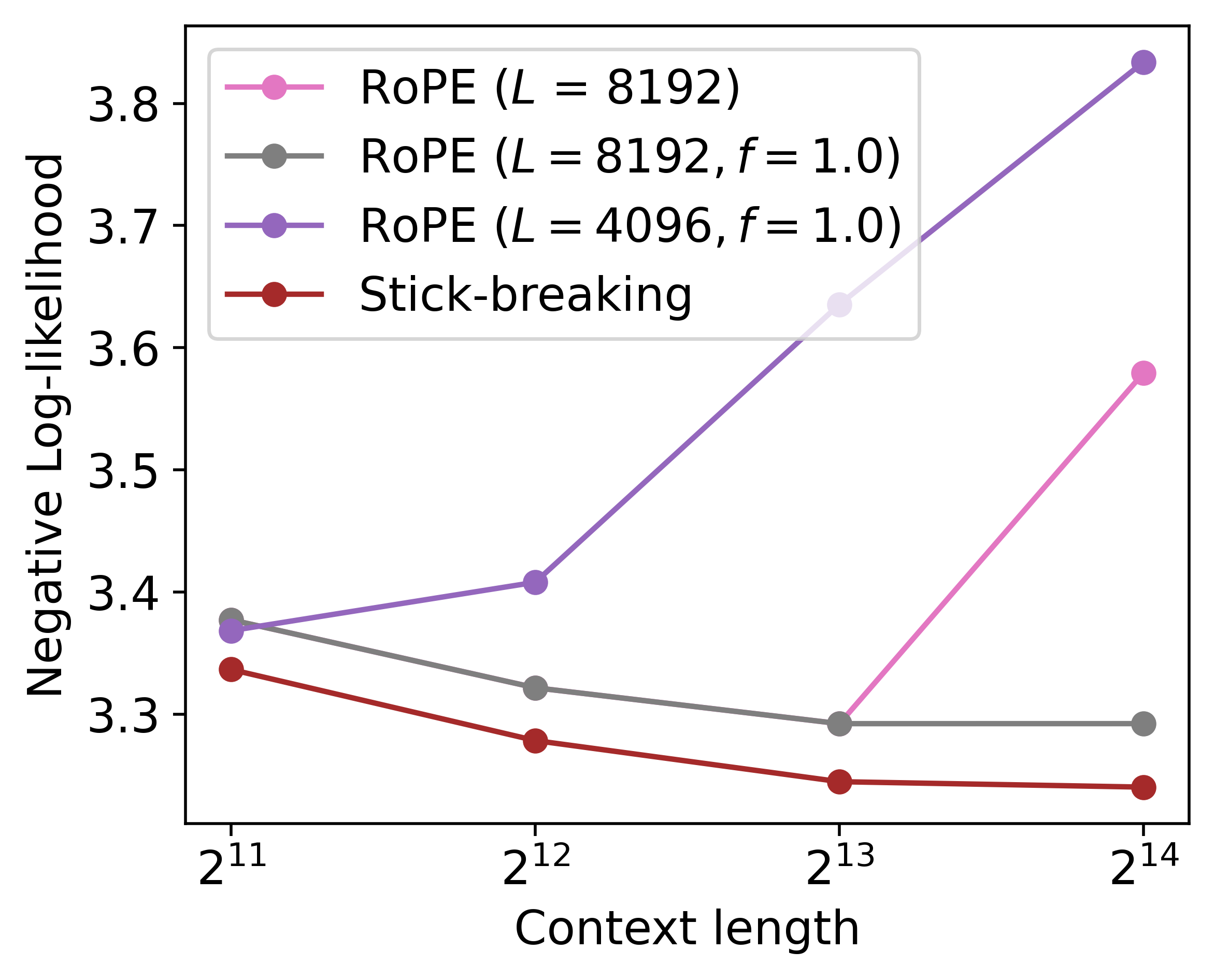}
    \caption{Comparison with $L=8192$}\label{fig:controllength}
\end{subfigure}
\begin{subfigure}{0.34\linewidth}
    \includegraphics[width=\textwidth]{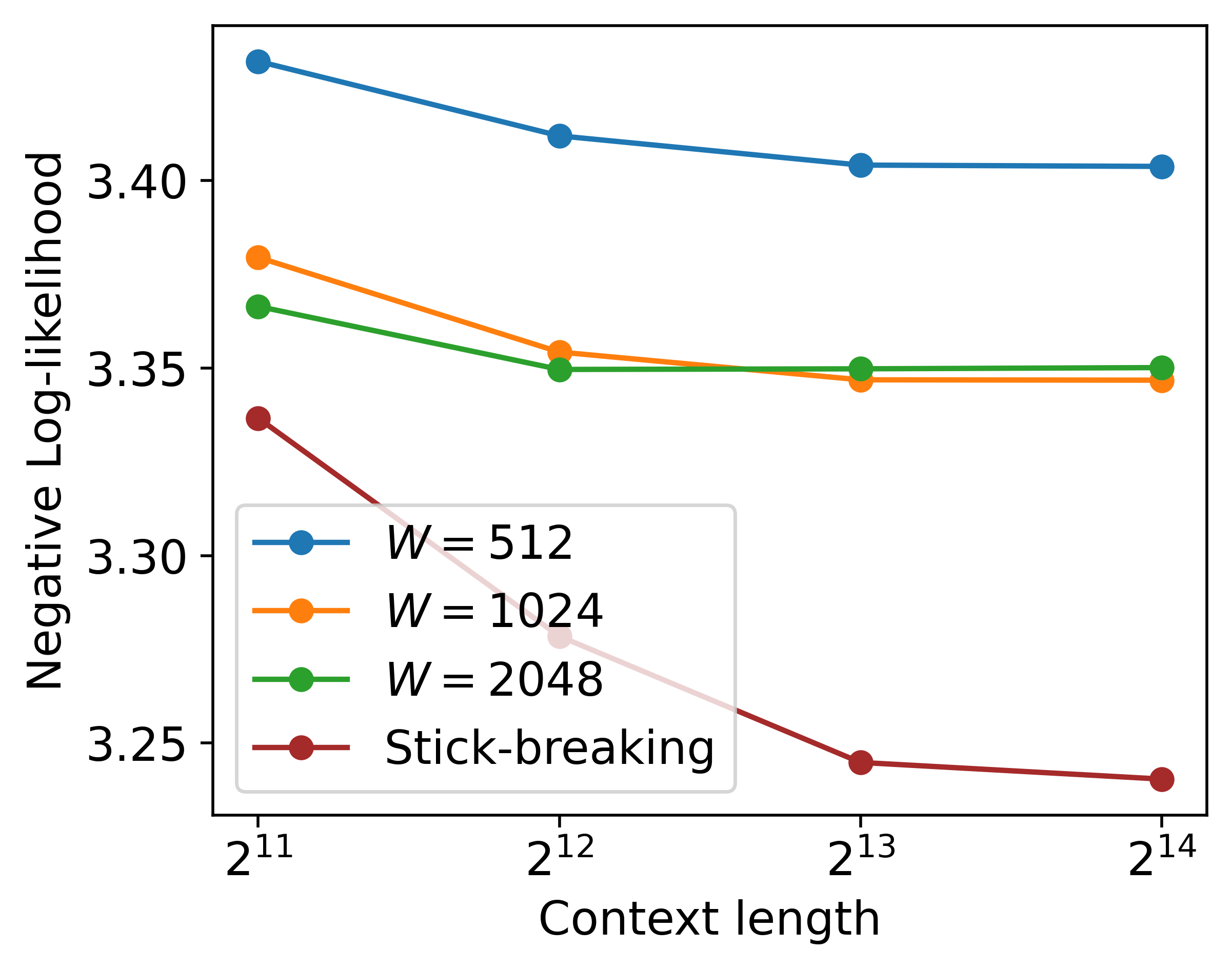}
    \caption{Sliding window}\label{fig:slidinglength}
\end{subfigure}
    \caption{Comparisons against different methods of sequence length extension. $L$ represents the training context length, $f$ is the RoPE scaling factor used, and $W$ is the window size in sliding window attention. 
    We compare against different position embeddings and biases with $L=8192$, training with $L=8192$, and various sliding window sizes with $L=2048$. 
    Note that the scale on the $y$-axis are different in all three plots.}
    \label{fig:lengen}
\end{figure}

We test the ability of stick-breaking for length generalisation, where we train on a fixed context length ($L=1024$) and test it on longer context lengths.
We started with the LLaMa 2 \citep{touvron2023llama} architecture, and modified the attention module to use the various baselines we compare against:
ALiBi \citep{press2021train}, FIRE \citep{li2023functional}, NoPE \citep{kazemnejad2024impact} and the default Llama position embedding RoPE \citep{su2021roformer}.
We trained on the first 15B tokens of SlimPajama \citep{cerebras2023slimpajama}, and evaluate it on the Wikitext benchmark in the LM evaluation harness \citep{eval-harness} , with context lengths of 2048 to 64K.
As control, we also trained a model with a context of $L=8192$, and as expected, the performance on longer sequences was better when RoPE scaling is used (See Figure \ref{fig:controllength}). 

Intuitively, we should expect a model that generalises well on longer sequences to have better likelihood as the context length is increased --- longer contexts mean more information for predicting the next word.
However, most methods do not generalise well to longer contexts (See Figure \ref{fig:length_extrapolation}). 

Contrary to the results in \citet{kazemnejad2024impact}, NoPE loss increases as context lengths are extended.
We also find that RoPE embeddings alone do not generalise, despite being a relative position encoding
Using RoPE scaling with the scaling factor $f = 1.0$ does alleviate the issue, but still results in an increase in the loss.
Surprisingly, while ALiBi is a fixed linear bias that increases with relative distance, it performs better than FIRE, which learns a function of the bias for a given relative distance. 
For position embeddings, our experiments suggest that RoPE with scaling works best for length extrapolation.
ALiBi is an incremental bias for the logits, effectively down-weighting positions further away, eventually approaching 0.
This can be viewed as a `soft' windowed attention, and given the relatively good performance, we also trained 3 models with windowed attention of $W \in \{512,1024, 2048\}$.
The results suggest that windowed attention helps with preventing the loss from spiking when the context length is extended. 
As expected, $W=512$ performs the worst.
However, $W=2048$ increases slightly as the context length is increased, while $W=1024$ decreases till $L=2^{13}$ and stays constant.
The $W=1024$ model is trained on context lengths of $L=2048$, which would allow the model to learn to deal with context lengths longer than the window size, while $W=2048$ is trained similar to the standard non-window attention model.
Figure \ref{fig:slidinglength} shows the generalisation curves.
Note that the $y$-axis in the plot is on a smaller scale than that of Figure \ref{fig:length_extrapolation}, indicating that the sliding window method is a relatively good method for length extrapolation as well.
In all cases, stick-breaking negative log-likelihood still decreases as the context length increases, outperforming the other methods.

\subsection{Language Model Pretraining}
\begin{table}[b]
    \centering
    \caption{
    Results on the various NLP benchmarks for the 1B and 3B pretrained model.
    `Softmax' benchmark is the standard Softmax + RoPE setting.}\label{tab:1B_results}
\resizebox{\linewidth}{!}{%
\begin{tabular}{lrrrrrrrrrr}
\toprule
\multicolumn{1}{r}{\textbf{Task}} & \textbf{ARC-c} & \textbf{ARC-e} & \textbf{Hella.} & \textbf{OBQA} & \textbf{PIQA} & \textbf{RACE} & \textbf{SciQ} & \textbf{Wino.} & \multirow{2}{*}{\textbf{Avg.}} & \textbf{Wiki.} \\

\multicolumn{1}{c}{} & \multicolumn{4}{c}{\textit{Accuracy (normalised)}} & \multicolumn{4}{c}{\textit{Accuracy}} &  & \multicolumn{1}{c}{\textit{Ppl.}} \\
\cmidrule(r){1-1} \cmidrule(lr){2-5} \cmidrule(lr){6-9} \cmidrule(lr){10-10}\cmidrule(l){11-11} 
\multicolumn{11}{l}{~\textit{1B Parameter Models}} \\[.3em]
Softmax & 35.8 & 65.6 & 64.8 & \textbf{38.8} & 75.0 & 36.5 & 90.5 & \textbf{63.4} & 58.8 & 13.8 \\
Stick-breaking & \textbf{37.7} & \textbf{67.6} & \textbf{65.4} & 36.6 & \textbf{76.0} & \textbf{37.4} & \textbf{91.9} & 63.1 & \textbf{59.5} & \textbf{13.4} \\
\midrule
\multicolumn{11}{l}{~\textit{3B Parameter Models}} \\[.3em]
Softmax & 42.2 & 73.1 & 73.2 & \textbf{40.8} & 78.8 & 37.4 & 93.5 & 67.6 & 63.3 & 11.3 \\
Stick-breaking & \textbf{44.9} & \textbf{74.3} & \textbf{74.1} & 40.4 & \textbf{79.7} & \textbf{37.8} & \textbf{93.9} & \textbf{68.0} & \textbf{64.1} & \textbf{10.8} \\
\cmidrule(lr){1-11}
Gemma2-2B & 50.0 & 80.2 & 72.9 & 41.8 & 79.2 & 37.3 & 95.8 & 68.8 & 65.8 & 13.1 \\
Qwen1.5-4B & 39.6 & 61.5 & 71.4 & 40.0 & 77.0 & 38.2 & 90.0 & 68.1 & 60.7 & 12.5\\
\bottomrule
\end{tabular}%
}
\end{table}

We pretrain the 1B and 3B models in this section using a two-stage training scheme in \citet{hu2024minicpm} and the Power learning rate schedule \citep{shen2024power}.
In the first stage, there is a warmup for the learning rate to 0.01, then we apply Power decay. 
Our training corpus has 1T tokens and mixes large-scale open-source datasets of medium quality with permissive licenses.
In the second stage, we exponentially decay the learning rate to zero. 
The stage 2 training corpus is a mix of stage 1 data and a small amount of high-quality open-source and synthetic corpora with permissive licenses.
The training batch size is 1024 and uses padding-free sequence packing for training in Dolomite Engine \citep{Mishra_Dolomite_Engine_A_2024}.
We evaluate the pretrained models on language model tasks and multiple-choice tasks from LM evaluation Harness~\citep{eval-harness}.
The multiple-choice tasks include:
grade-school science questions (ARC;~\citealt{clark2018think}),
common sense reasoning (Hellaswag;~\citealt{zellers2019hellaswag}),
open book question answering (OpenBookQA;~\citealt{mihaylov2018can}),
physical questions (PIQA;~\citealt{bisk2020piqa}), 
reading comprehension (RACE; \citealt{lai2017large}), and
Winograd schema task (Winogrande;~\citealt{sakaguchi2021winogrande}).
Table~\ref{tab:1B_results} shows the performance. 
Overall, we outperform our own pretrained standard attention models that are trained on the same settings. 
We perform better on average, and attain better perplexity on Wikitext.

\begin{wraptable}{r}{0.45\linewidth}
    \centering
        \caption{MMLU few-shot results}\label{tab:1B_mmlu_results}
\begin{tabular}{lrr}
\toprule
 & \multicolumn{2}{c}{\textbf{MMLU}} \\
\multicolumn{1}{r}{} & 0-shot & 5-shot \\
\midrule
\multicolumn{3}{l}{\textit{1B Parameter Model}} \\[.2em]
Softmax & 25.7 & 25.2 \\
Stick-breaking & \textbf{28.4} & \textbf{29.3} \\
\cmidrule(lr){1-3}
TinyLlama & 25.3 & 26.0 \\
\midrule
\multicolumn{3}{l}{\textit{3B Parameter Model}} \\[.2em]
Softmax & 46.1 & 49.1 \\
Stick-breaking & \textbf{50.8} & \textbf{52.9} \\
\cmidrule(lr){1-3}
Gemma2-2B  & 49.3 & 53.1 \\ 
Qwen1.5-4B & 54.2 & 55.2 \\
\bottomrule
\end{tabular}
\vspace{.5em}

\caption{3B Model GSM8K Results}\label{tab:3b_gsm8k_results}
\begin{tabular}{lrr}
\toprule
 & \multicolumn{2}{c}{\textbf{GSM8K}} \\
\multicolumn{1}{r}{} & 5-shot & 8-shot, CoT \\
\midrule
Softmax & \textbf{44.1}  & 44.2 \\
Stick-breaking & 42.3 & \textbf{49.7} \\
\bottomrule
\end{tabular}
\end{wraptable}
We also evaluated the models on MMLU with 0-shot and 5-shot settings.
For the 1B models, we have included the results for TinyLlama~\citep{zhang2024tinyllama}
Due to the inductive bias of stick-breaking, we believed that stick-breaking would perform better on MMLU in a few-shot setting as it would not be distracted by the few-shot examples provided in the context. 

Stick-breaking performs surprisingly well even in the 0-shot setting, and improves with few-shot examples provided.
Note that this may not always be the case, as in our Softmax + RoPE model, the performance decreases with few-shot examples in context.

We have included the Qwen1.5-4B and Gemma2-2B models for comparison, and our 3B model underperforms Gemma2-2B, a smaller model, while it outperforms Qwen1.5-4B.

Finally, we evaluate our 3B model on the GSM8K dataset \citep{cobbe2021training}.
Interestingly, the 5-shot setting underperforms standard attention while CoT with 8-shot sees an improvement of 5.5\%.
The improvements in MMLU and GSM8K on a few-shot setting may suggest stick-breaking has a better inductive bias for in-context learning, particularly in the formats that these evaluation benchmarks use.

\subsubsection{1B Length extrapolation}
\begin{figure}
  \centering
\begin{subfigure}{0.3295\linewidth}
    \includegraphics[width=\linewidth]{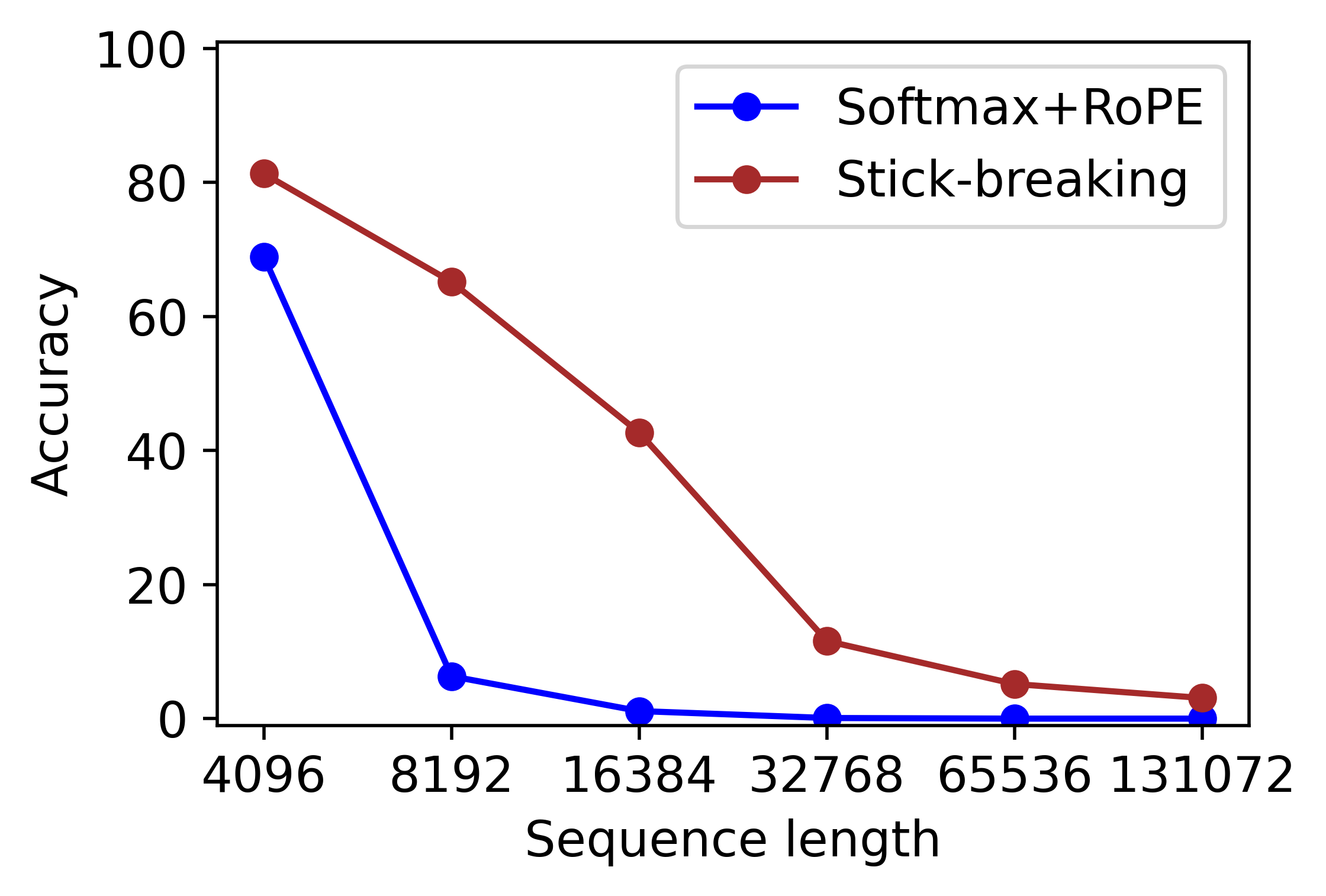}
    \caption{Overall}\label{fig:ruler_overall}
\end{subfigure}
\begin{subfigure}{0.3295\linewidth}
    \includegraphics[width=\linewidth]{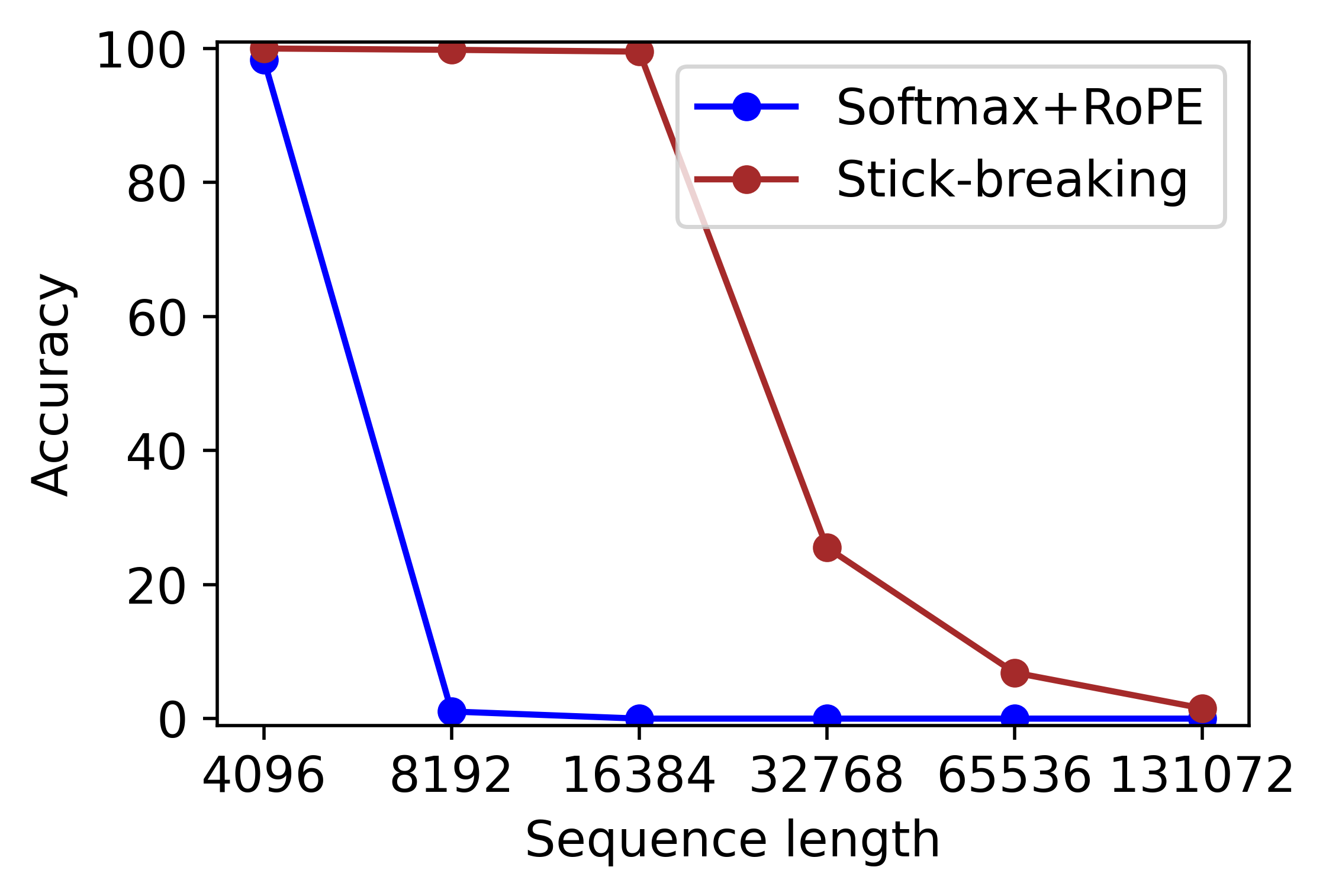}
    \caption{Needle in a Haystack (NIAH)}\label{fig:ruler_niah}
\end{subfigure}
\begin{subfigure}{0.3295\linewidth}
    \includegraphics[width=\linewidth]{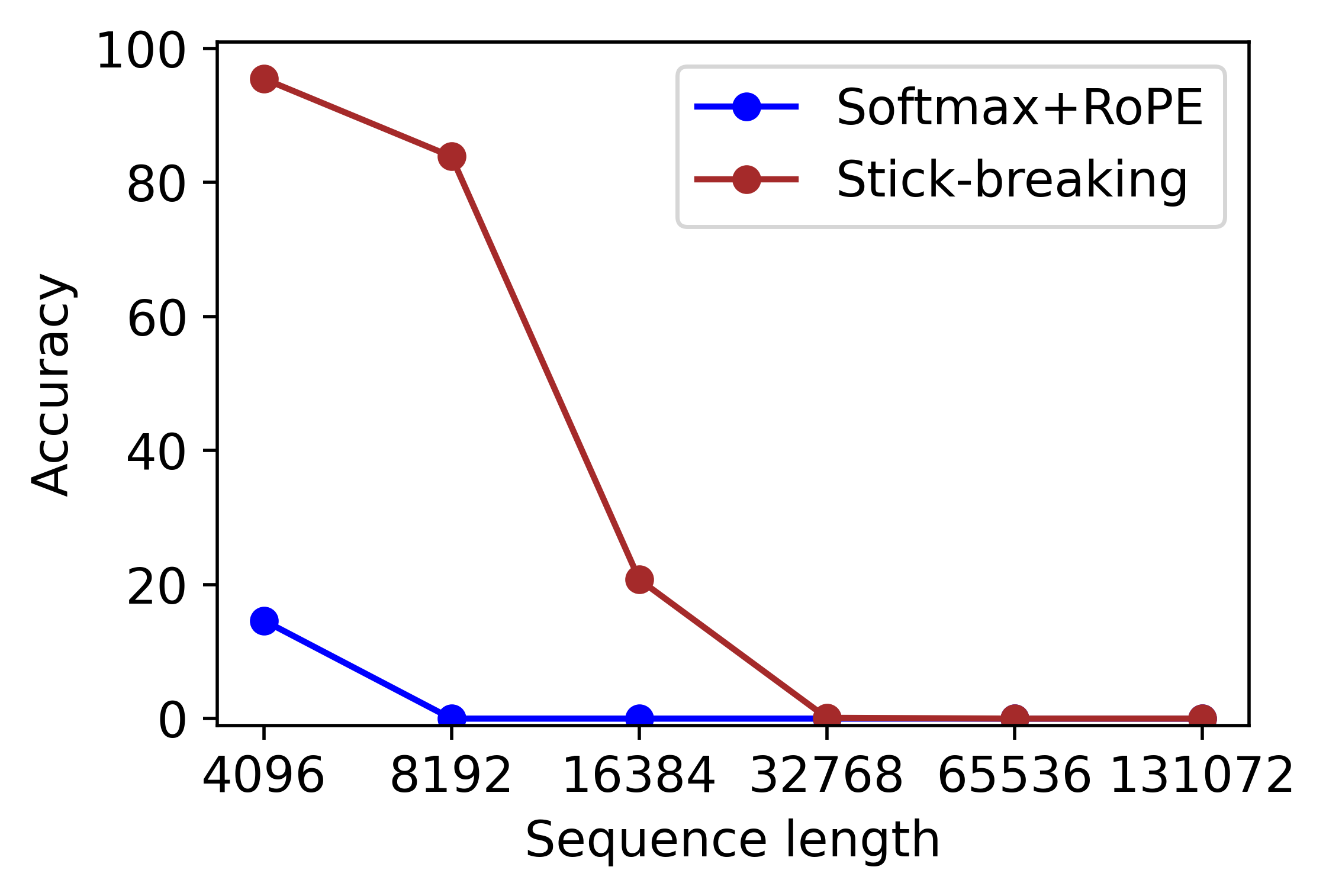}
    \caption{Variable Tracking}\label{fig:ruler_vt}
\end{subfigure}
\caption{Results on the RULER benchmark. Without further finetuning for long context, we evaluate the 1B models on the RULER tasks. Subfigures: \ref{fig:ruler_overall} shows the average over all 13 tasks,  \ref{fig:ruler_niah} is the average across the various NIAH tasks,  \ref{fig:ruler_vt} is the result for the variable tracking task.}
\end{figure}
We test the 1B stick-breaking and softmax model with the RULER benchmark \citep{hsieh2024ruler}.
The benchmark consists of `needle-in-a-haystack'-like tasks, and is generally used for testing retrieval capabilities of long-context models that are trained specifically for long contexts. 
In our setting, we use RULER to evaluate both 1B models trained on 4096 contexts. 
Accordingly, the general capabilities of these models on longer contexts are much worse than purpose-trained models.

On average, the performance of stick-breaking dominates Softmax + RoPE with scaling (Figure \ref{fig:ruler_overall}).
In the breakdown, we find that the stick-breaking model is surprisingly good at extrapolating on NIAH tasks up to 16K context lengths, while the standard model significantly drops in performance.
Our results on the variable tracking (VT) task agrees with our experiments on MQRAR.
The task involves tracking the variable assignments provided in the context, and we find that even in-distribution ($L=4096$), the standard model does not perform well at this task. 

\section{Conclusion \& Future Work}
We propose a formulation for using the stick-breaking process as a replacement for softmax for attention.
Stick-breaking attention allows us to do away with position embeddings, while still retaining model performance.
We detail the specifics of implementing the stick-breaking kernel in Triton for large scale training.
We then demonstrate that stick-breaking is good at length extrapolation, performing better than other position embedding and position bias methods in our 350M class models.
We also show that our pretrained stick-breaking models peform better in a controlled experiment, given the same training data and training regime.
On retrieval in the RULER benchmark, stick-breaking outperforms softmax attention.

The drawbacks in efficiency can be improved in future work by making similar optimisations that Flash Attention \citep{dao2022flashattention,dao2023flashattention} made for speedups. 
These include making full use of features in CUDA, and by cache retrieval optimisations.
We believe there is plenty of room for improvement in computation efficiency that can be made in future versions of stick-breaking.
We also think there can be stick-breaking-specific modifications to Transformers that can augment the Stick-breaking Transformer, and we have initial results in Appendix \ref{sec:stickbreaking_transformer}.
Overall, we find stick-breaking attention to be a promising replacement for Softmax + RoPE in Transformer models.

\ificlrfinal
\subsubsection*{Acknowledgments}
We would like to thank Mayank Mishra and Gaoyuan Zhang for their help during the training of the 1B and 3B models.
\fi


\bibliography{iclr2025_conference}
\bibliographystyle{iclr2025_conference}
\newpage
\appendix
\section{Derivation of Log-space Formulation}\label{eqn:logspacederiv}
$\sigma(x)$ is the sigmoid function,
\begin{align}
    \sigma(x) &= \frac{1}{1 + \exp(-x)} = \frac{\exp(x)}{1 + \exp(x)} \\
    1 - \sigma(x) &=  \frac{1}{1 + \exp(x)}
\end{align}
Since $\beta_{i,j} = \sigma(z_{i,j})$,
\begin{align}
 \log \beta_{i,j} &= z_{i,j} -\log \left(1 + \exp(z_{i,j})\right) \\
    \log \left(1 - \beta_{i,j}\right) &= \log \frac{1}{1 + \exp(z_{i,j})} =  -\log \left(1 + \exp(z_{i,j})\right)
\end{align}
We can substitute these back in,
\begin{align}
    \mA_{i,j} &= \exp \left( \log \beta_{i,j} + \sum_{k=i+1}^{j-1} \log \left( 1 - \beta_{k,j} \right)   \right) \\
    &= \exp\left( z_{i,j} -\log \left(1 + \exp(z_{i,j})\right) - \sum_{k=i+1}^{j-1}  \log \left(1 + \exp(z_{k,j})\right) \right) \\
    &= \exp\left( z_{i,j} - \sum_{k=i}^{j-1}  \log \left(1 + \exp(z_{k,j})\right)   \right)
\end{align}
\newpage
\newpage

\section{1B Model RULER Results}
\begin{figure}[H]
    \centering
    \includegraphics[width=\linewidth]{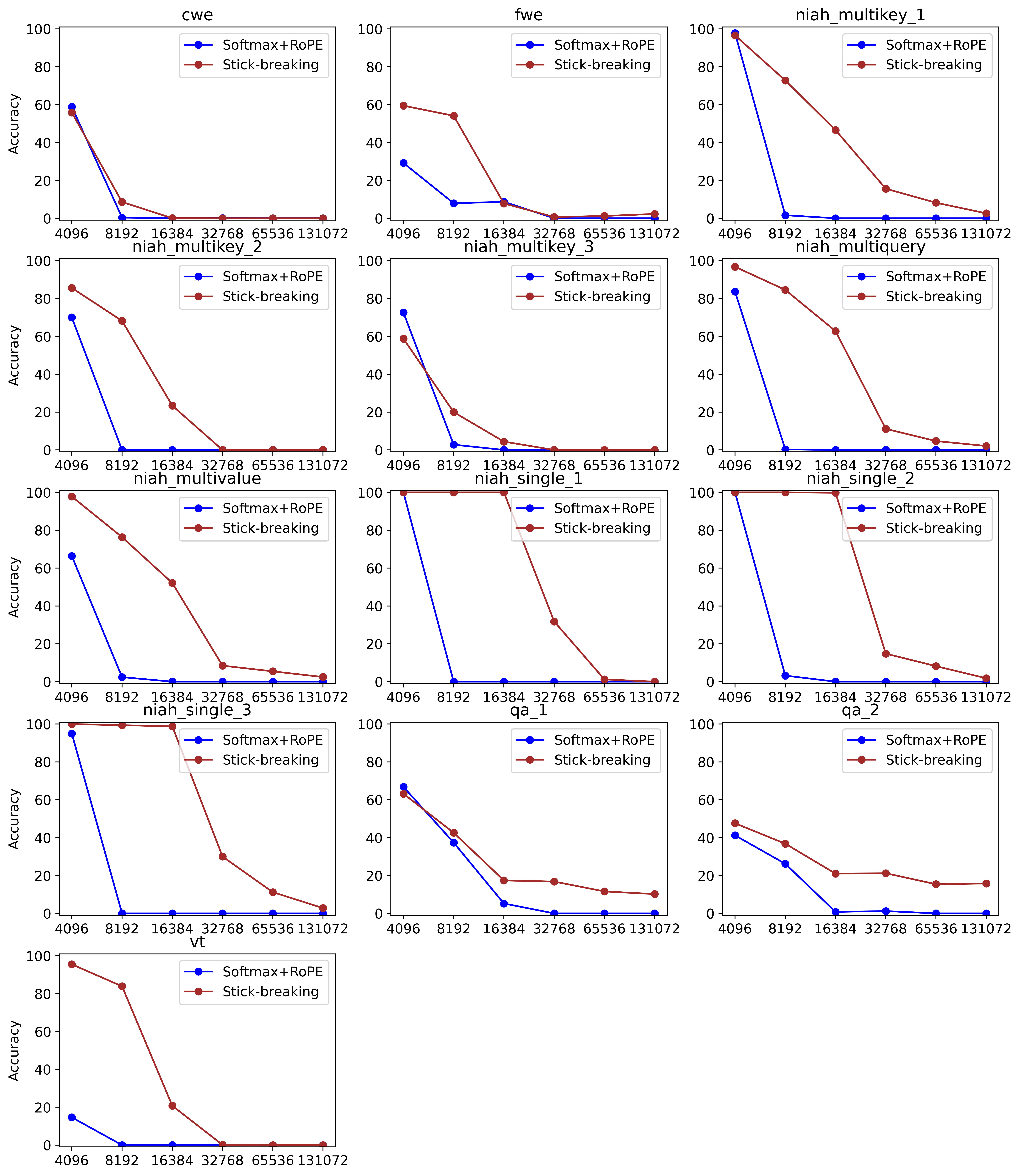}
    \caption{RULER length extrapolation results}
    \label{fig:enter-label}
\end{figure}
\newpage
\section{Stick-breaking Specific Transformer Modifications}\label{sec:stickbreaking_transformer}
While we demonstrate the effectiveness of stick-breaking as a drop-in replacement for softmax attention in the main paper, found several minor modifications to the standard Transformer architecture can help enhance the performance of the Stick-breaking Transformer on various tasks. 
\begin{table}[]
\centering
\small
\caption{In these experiments, we use $n_\mathrm{layer} = 40$, uses Grouped Query Attention with $n_\mathrm{head}=12$ and 4 key and value heads, and $d_\mathrm{head} = 128$.
For the 6.5B model, we use $n_\mathrm{embd} = 1536$, with 64 experts and $k=8$, with each expert having a hidden size of 512.
For the 28.6B model, we use $n_\mathrm{embd} = 4096$, with 72 experts and $k=12$, with each expert having a hidden size of 768.}\label{tab:large_model_results}
\scalebox{0.95}{
\begin{tabular}{@{}lllrrrrrrrrr@{}}
\toprule
& \multicolumn{1}{r}{\textbf{RB}} & \multicolumn{1}{r}{\textbf{GN}} & \textbf{ARC} & \textbf{Hella.} & \textbf{OBQA} & \textbf{PIQA} & \textbf{Wino.} & \textbf{MMLU} & \textbf{GSM8K} & \textbf{MATH} & \multirow{2}{*}{\textbf{Avg.}} \\
& & & \multicolumn{3}{c}{\textit{Accuracy (normalised)}} & \multicolumn{3}{c}{\textit{Accuracy}} & \textit{8-shot, cot} & \textit{4-shot} &  \\
\cmidrule(r){1-3} \cmidrule(lr){4-6} \cmidrule(lr){7-9} \cmidrule(lr){10-10} \cmidrule(lr){11-11} \cmidrule(l){12-12} 
\multicolumn{12}{l}{\textit{1.2B parameters, 1.25T tokens}} \\[.2em]
Softmax & -- & -- & 55.8 & 66.7 & 38.4 & 76.0& 63.9 & 28.3 & 27.0 & 7.6 & 45.5 \\
SB & \xmark & \xmark & 54.0 & 66.9 & 39.6 & 76.8 & 65.5 & 37.3 & 29.6 & 8.2 & 47.2 \\
& \cmark & \xmark & 54.1 & 66.6 & 39.8 & 76.2& 66.9 & 39.6 & 29.2 & 7.8 & 47.5 \\
& \cmark & \cmark & 56.4 & 66.7 & 38.4 & 76.7& 64.8 & 39.8 & 29.3 & 8.2 & 47.5 \\
\cmidrule(r){1-3} \cmidrule(lr){4-6} \cmidrule(lr){7-9} \cmidrule(lr){10-10} \cmidrule(lr){11-11} \cmidrule(l){12-12} 
\multicolumn{12}{l}{\textit{1B activation / 6.5B parameters, 2T tokens}} \\[.2em]
Softmax & -- & -- & 66.3 & 76.9 & 44.6 & 80.7 & 72.5 & 55.5 & 50.8 & 21.6 & 58.6 \\
SB & \cmark & \cmark & 68.7 & 77.9 & 46.6 & 80.5 & 74.4 & 56.9 & 55.3 & 22.5 & 60.4 \\
\cmidrule(r){1-3} \cmidrule(lr){4-6} \cmidrule(lr){7-9} \cmidrule(lr){10-10} \cmidrule(lr){11-11} \cmidrule(l){12-12} 
\multicolumn{12}{l}{\textit{6B activation / 28.6B parameters, 3T tokens}} \\[.2em]
Softmax & -- & -- & 76.8 & 84.4 & 51.2 & 84.1 & 80.1 & 67.7 & 70.4 & 34.5 & 68.6 \\
SB & \cmark & \cmark & 77.2 & 84.7 & 51.4 & 83.7 & 81.5 & 71.2 & 71.9 & 35.6 & 69.6 \\
\bottomrule
\end{tabular}}
\end{table}

\paragraph{Remainder Bias (RB)}
As $ \sum_{i=1}^{j-1} \mA_{i,j} \leq 1$,   we experimented with using remaining attention weight is assigned to an embedding $\vr$, by modifying the output of the attention layer as follows:
\begin{align} 
      \vo_{j} &= \sum_{i=1}^{j-1} \mA_{i,j} \cdot \vv_i + \left(1 - \sum_{i=1}^{j-1} \mA_{i,j}\right) \cdot \vr \label{eqn:rem_output}
\end{align}
Each head has its corresponding $\vr$ embedding, resulting in a matrix of additional parameters of $n_\mathrm{head} \cdot d_\mathrm{head}$.
We refer to this as remainder bias.
This is similar to an attention sink token \citep{xiao2023efficient}.
The authors showed that having an attention sink token of $\mathbf{0}$ resulted in poorer performance than the standard attention, and saw better performance when a learnable attention sink embedding was used.
In our experiments, we saw improvements when the remainder bias was used. 

\paragraph{Group Norm (GN)}
Due to the unnormalised weight of stick-breaking attention, the resulting $\vo_j$ can have varying norm magnitudes.
We experimented with applying a head-wise norm, which was implemented with Group Norm.
This was also done with Differential Transformers \citep{ye2024differential}, for similar reasons.
We found improvements when Group Norm was used in conjunction with RB. 

In Table \ref{tab:large_model_results}, we perform experiments with RB and GN on 1B models where we found improvements on 1B dense models.
We then trained a 6.5B and 28.6B parameter MoE model with both GN and RB, and show that they outperform a corresponding Transformer with softmax attention. 

\subsection{Context window extension}
All models are pre-trained with a context window of 4096.
To extend their context lengths, we continue training for 6250 steps at a context length of 16k. 
Our learning rate is scheduled to increase from $10^{-5}$ to $0.000125$ in 150 steps, and then decays exponentially to 0 until 6250 steps.
The effective batch size has 4M tokens.

For Softmax+RoPE, we set the dynamic RoPE scaling factor to 4.
In Figure \label{fig:length-extension}, we show that by applying rope scaling, the model can somewhat extrapolate to 16k context lengths, but the log-likelihood spikes as we extend it further.

Applying this method of context window extension to both models, we find that the Softmax+RoPE model improves slightly, while still showing issues with longer context lengths, albeit mitigated slightly.
Stick-breaking however exhibits lower NLL as we extend the context window further to 32k and 64k.

The behaviour of RoPE out-of-the-box for length extrapolation is also demonstrated in Section \ref{sec:350extrapolation}. 
We also show that training at longer sequences does not mitigate the out-of-distribution performance of Softmax+RoPE models.
Here, we show that while naive continued fine-tuning helps, the improvement on long-context windows is not as substantial as with stick-breaking.
This suggests better context window extension behaviour and further length extrapolation for the stick-breaking attention mechanism.

\begin{figure}[H]
    \centering
    \includegraphics[width=0.75\linewidth]{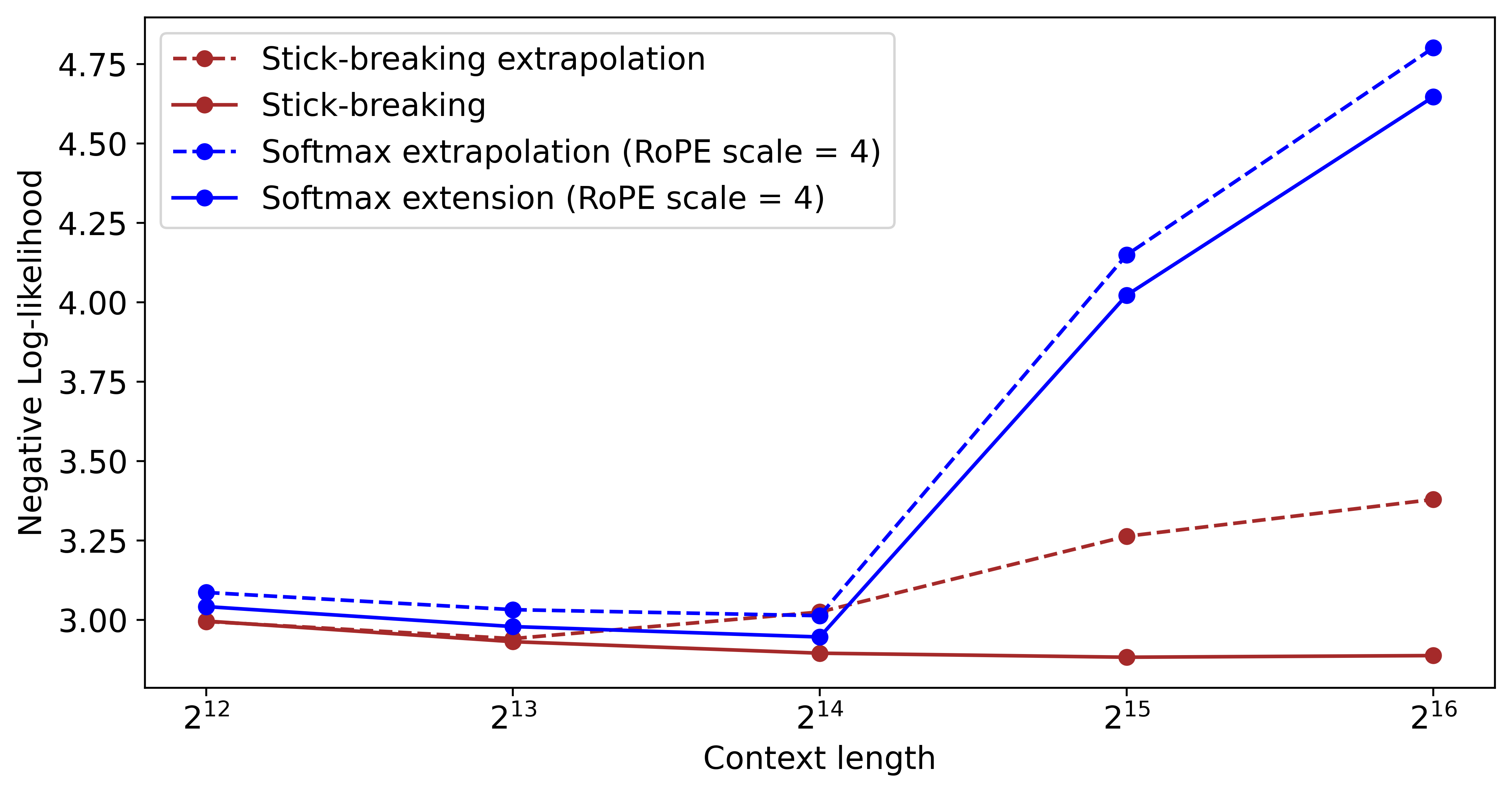}
    \caption{Stick-breaking and Softmax+RoPE context window extrapolation performance on longer context lengths in the Pile 10k evaluation dataset.}
    \label{fig:length-extension}
\end{figure}

\end{document}